# A machine-learning-assisted progressive digit-randomness screening framework for detecting non-random patterns in raw numerical research data

Zhuohua Cao[a,*]

[a] Key Laboratory of Natural Medicines of the Changbai Mountain, Ministry of Education, College of Pharmacy, Yanbian University, Yanji 133002, Jilin Province, China

[*] **Correspondence:**

Zhuohua Cao, M.Med. Key Laboratory of Natural Medicines of the Changbai Mountain, Ministry of Education, College of Pharmacy, Yanbian University, Yanji 133002, Jilin Province, China. E-mail address: 1247166997cc@gmail.com

# Abstract

Raw numerical datasets remain less systematically examined in integrity screening than images, plagiarism, or summary-statistic inconsistencies. We developed the Fabrication-risk Digit Randomness Screening model (FDRS), a statistical and machine-learning framework for detecting non-random digit-pattern irregularities in numerical research data. FDRS integrates single- and joint-decimal-digit tests, Cramer's V, entropy metrics, Kullback-Leibler divergence, digit-preference indices, progressive subsampling, and semi-supervised risk scoring. It was evaluated using an instrument-derived enzymatic absorbance dataset (RawData, n=253) and a blinded manually simulated irregular dataset (ErrData, n=255). RawData showed no significant deviation in single third-decimal-digit analysis, whereas ErrData showed a significant deviation. In joint third-fourth decimal digit analysis, ErrData showed higher Cramer's *V*, lower normalized entropy, higher KL divergence, and a more persistent progressive-subsampling deviation signal. In internal validation, Elastic-net Logistic Regression achieved the highest AUC (0.98395) and lowest Brier score (0.048439), while Random Forest achieved the highest accuracy (0.926667) and balanced accuracy (0.935). RawData received a low ensemble risk score of 0.124627 and was classified as Grade 0; ErrData received a score of 0.740760 and was classified as Grade 3. External real-world benchmarks supported graded risk stratification: three datasets without identified public post-publication concerns were classified as Grade 0 or 1, whereas two datasets from publicly questioned or institutionally handled articles were classified as Grade 2 or 3. FDRS can prioritize raw numerical datasets for further review by integrating interpretable statistical and machine-learning features. It is an auxiliary digit-structure screening tool, not standalone evidence of fabrication or misconduct.

## 1 Introduction

Research data integrity is a foundational requirement for scientific reliability, reproducibility, and public trust[1, 2]. Fabrication, falsification, and selective modification of data are among the most serious forms of research misconduct. Although their exact prevalence is difficult to estimate, survey-based meta-analyses suggest that a non-negligible proportion of researchers have admitted to at least one instance of fabrication, falsification, or modification of data or results, while substantially higher proportions report having observed such behaviors among colleagues. Fanelli's meta-analysis estimated that approximately 1.97% of scientists admitted to fabricating, falsifying, or modifying data or results at least once[3], whereas an updated meta-analysis by Xie et al. reported a pooled estimate of 2.9% for at least one form of fabrication, falsification, or plagiarism[4]. Accordingly, journal editors, institutions, and research integrity bodies have increasingly emphasized systematic procedures for handling suspected fabricated data, as reflected in publication-ethics guidance from the Committee on Publication Ethics[5, 6].

Current research-integrity screening practices have focused heavily on image manipulation, plagiarism, duplicate publication, and inconsistencies in reported statistical summaries[7-9]. However, many experimental results are ultimately recorded as numerical raw data, including absorbance values, cycle threshold values, fluorescence intensities, cell counts, densitometric measurements, biochemical readouts, and pharmacological response values. Numerical datasets can be affected not only by deliberate fabrication but also by selective editing, excessive rounding, copy-and-paste reuse, transcription error, or inappropriate post hoc adjustment. In high-throughput biomedical research, forensic bioinformatics has shown that errors in raw data structure, annotation, or processing can seriously compromise reproducibility and downstream conclusions[10]. Therefore, statistical tools designed specifically for numerical raw data may provide an important complement to image-based and text-based integrity screening.

A classical approach to numerical irregularity detection is based on digit-distribution analysis. Benford's law and related first-digit tests have been used in accounting, social sciences, elections, and scientific-data forensics[11]. Previous work has developed machine learning models for detecting data fraud in the biomedical field based on Benford's Law[12]. However, first-digit approaches are not universally appropriate for laboratory datasets, particularly when values occupy a narrow numerical range, are constrained by measurement instruments, or are reported with fixed decimal precision. In contrast, terminal or non-leading digits may carry information about whether the fine-scale numerical structure is consistent with a plausible measurement process. Mosimann, Wiseman, and Edelman demonstrated that even when individuals consciously attempt to generate random digits, human-generated numbers may deviate from uniform digit distributions[13, 14]. The U.S. Office of Research Integrity has also described statistical-forensic reasoning in which rightmost or otherwise inconsequential digits may reveal personal preferences in questioned data, particularly when those digits should be approximately uniform under the relevant data-generating mechanism[15, 16].

Previous studies have applied statistical approaches to identify suspicious numerical patterns in clinical or scientific datasets. Al-Marzouki et al. evaluated statistical methods for detecting possible data fabrication in clinical trials by examining baseline comparisons, digit preference, and distributional patterns[17]. Carlisle further applied statistical tests to thousands of randomized controlled trials and showed that some studies exhibited distributions inconsistent with random sampling expectations[18]. A scoping review of methods for assessing research misconduct in health-related research identified multiple approaches for data concerns but also noted that many available methods are example-based, insufficiently standardized, and often lack formal validation[19]. These limitations highlight the need for transparent, reproducible, and interpretable models that can screen numerical raw data without overstating their evidentiary meaning.

Based on this rationale, we developed the Fabrication-risk Digit Randomness Screening model (FDRS) as a statistical and machine-learning-assisted framework for identifying non-random digit patterns in raw numerical datasets. The central assumption of FDRS is that, under a defined measurement mechanism and sufficient numerical precision, later decimal digits of independent continuous measurements should follow an approximately random or empirically referenceable pattern, whereas manually fabricated or modified values may carry subjective digit preferences. The preliminary FDRS framework proposed single-decimal-digit multinomial testing, multi-decimal joint distribution testing, and progressive subsampling analysis as core statistical modules. In the single-digit model, the $j$-th decimal digit $D_{i\,j} \in \{0,1,\ldots,9\}$ is tested against an expected uniform distribution; in the joint-digit model, adjacent decimal combinations such as $D_{i\,3,4} \in \{00,01,\ldots,99\}$ are evaluated to detect combination-level irregularities.

A key feature of FDRS is the integration of conventional statistical testing with progressive subsampling. Single full-sample tests may be unstable: small samples may have low statistical power, whereas large samples may produce statistically significant but practically trivial deviations. Therefore, FDRS evaluates whether digit-distribution deviations remain stable across gradually increasing sample sizes. Stable elevation of Cramér's $V$ reduced standardized entropy, increased Kullback-Leibler divergence, or persistent enrichment of specific terminal-digit combinations may provide stronger evidence for a non-random digit-generating process than a single $P$-value alone. In the preliminary design, FDRS further extracted digit frequencies, two-digit and three-digit combination frequencies, chi-square statistics, Cramér's $V$, Shannon entropy, KL divergence, adjacent-digit dependence, repeated-tail ratios, 0/5 preference indices, odd-even ratios, high-low digit ratios, and progressive curve features as candidate predictors for downstream modeling.

To improve robustness and pattern-recognition capacity, we incorporated semi-supervised machine learning into the FDRS framework. Because confirmed fabricated datasets are rarely available and ethically difficult to obtain, the model uses a hybrid training strategy consisting of simulated normal datasets, digit-preference simulated datasets, and mixed-contamination datasets. Random Forest was used to capture nonlinear and interaction-based digit patterns, Elastic-net Logistic Regression provided sparse and interpretable risk modeling, support vector machines were applied to learn nonlinear decision boundaries, and Isolation Forest was used as an anomaly detector trained primarily on normal-reference patterns. These algorithms have well-established methodological foundations in statistical learning and anomaly detection.

Importantly, the proposed FDRS framework is not intended to prove fabrication or to replace institutional investigation, raw-record verification, or expert review. Digit irregularities may arise from benign sources such as instrument precision, rounding rules, bounded measurement ranges, spreadsheet formatting, batch processing, data transformation, or transcription practices. Thus, FDRS should be interpreted as a risk-screening and prioritization tool: it identifies datasets whose numerical digit structure deviates from an expected or reference pattern and therefore warrants further verification. This study aimed to formalize the FDRS framework, define its statistical and machine-learning components, and evaluate its feasibility for screening non-random digit patterns in raw numerical research data.

# 2 Materials and methods

## 2.1 Study design

This study aimed to develop a statistical–machine learning framework for screening non-random digit patterns in raw numerical research data, termed the FDRS. The model is based on the assumption that, under a defined data-generating mechanism, sufficient numerical precision, and absence of systematic rounding bias, the later decimal digits of continuous raw measurements should exhibit an approximately random distribution. In contrast, manually fabricated, selectively modified, copied, pasted, or excessively adjusted data may show detectable non-random patterns in decimal digit frequencies, multi-digit combinations, entropy, adjacent-digit dependence, and progressive subsampling stability.

The overall FDRS workflow consisted of six major steps: input of raw numerical data arranged in a single column; extraction of selected decimal digits; single decimal-digit multinomial testing and multi-decimal joint distribution testing; progressive subsampling analysis to evaluate the stability of digit-distribution deviations; feature extraction from digit-frequency, information-theoretic, psychological-preference, and progressive-curve metrics; and semi-supervised machine-learning modeling to generate a numerical risk score for non-random digit-pattern irregularity.

**2.2 Dataset construction**

2.2.1 Real raw dataset

The real dataset, referred to as RawData, consisted of enzymatic activity absorbance values measured in the laboratory using a SpectraMax 190 absorbance microplate reader (Molecular Devices, San Jose, CA, USA). Each absorbance value represented an independent raw numerical readout generated during enzyme-activity measurement. The raw values were exported and stored as a single-column txt file, with each row corresponding to one absorbance measurement. To ensure consistency in decimal digit extraction, the absorbance values were analyzed in their original numerical format whenever possible. Values with fewer decimal places than required for the target digit positions were right-padded with zeros during digit extraction. Values recorded in scientific notation, if present, were converted into fixed-decimal format before analysis. No transformation, normalization, or rounding beyond the original exported format was applied before FDRS analysis.

2.2.2 Blinded manually simulated irregular dataset

The manually simulated irregular dataset, referred to as ErrData, was generated by a volunteer under blinded conditions. The volunteer was asked to manually create a set of plausible enzyme-activity-like absorbance values with the same general numerical format and decimal precision as the real raw dataset. During this process, the volunteer was blinded to the FDRS algorithm, the specific digit-distribution features used by the model, the expected statistical tests, and the downstream machine-learning classification criteria. This dataset was designed to represent a pseudo-experimental numerical dataset generated through human manual input rather than instrumental measurement. It was used as a proof-of-concept irregular reference dataset to evaluate whether FDRS could detect non-random digit structures potentially introduced by human number-generation behavior. The ErrData dataset was not derived from actual experimental measurements and was not treated as evidence of real research misconduct. Its role was limited to methodological validation and assessment of model sensitivity to human-generated digit-pattern irregularities.

2.2.3 Benchmark real-world datasets

To evaluate the external performance of the proposed digit-randomness screening framework, five real-world numerical datasets were manually curated from published source data files and figure-associated raw data tables. These datasets were not used during model construction or parameter tuning and were used only as independent benchmark datasets for post hoc model evaluation.

The benchmark datasets were divided into two categories: RealRawData and RealErrData. The RealRawData datasets were defined as published numerical datasets for which, to the best of our knowledge at the time of analysis, no public post-publication concerns, formal investigations, or data-integrity actions had been identified for the specific analyzed numerical records. In contrast, the RealErrData datasets were defined as published numerical datasets selected from articles or figure panels that had been publicly questioned and for which relevant data-integrity concerns, corrections, or institutional/journal-level handling records were available to the authors[20-26]. Because the benchmark datasets differed in numerical format and reported decimal precision, the target decimal positions were selected according to the original data structure. The third-fourth decimal digit combinations were analyzed for RealRawData1 and RealRawData2, whereas the first-second decimal digit combinations were analyzed for RealRawData3, RealErrData1, and RealErrData2. This position-specific strategy was applied to preserve the original reporting precision of each dataset and to avoid imposing a uniform digit-position rule across heterogeneous numerical sources. This classification was used only for benchmark evaluation and should not be interpreted as an independent adjudication of research misconduct by the present model.

RealRawData1 was extracted from Fig. 1P of the article by Moro et al., specifically the EV siISG15 IdU/CIdU values. The article investigated the role of ISG15 in replication fork stability and cell viability in BRCA-defective cells[27]. RealRawData2 was extracted from Fig. 3B of the article by Guo et al., using the reported RT-qPCR Ct values associated with the SunCatcher clonal barcoding system[28]. RealRawData3 was extracted from Fig. 2h of the article by Khan et al., corresponding to tumor-volume measurements from SW620 shDKC1 and control cells subcutaneously implanted into NOD/SCID mice[29].

RealErrData1 was curated from the article by Zheng et al., including the raw numerical values from Fig. 3 for statistics of area per organoid stained with EthD-1 (%), flow-cytometry percentage values from Figs. 5 and 6, and the raw data associated with Supplementary Fig. S3B[20, 24, 25]. RealErrData2 was curated from the article by Jin et al., using all two-decimal-place numerical records available from Supplementary Figs. S4 and S6[21-23, 26]. The corresponding article was later accompanied by an Author Correction, although the present analysis used the dataset label only as an external benchmark category rather than as an independent misconduct determination.

All values were extracted as numerical records and stored as single-column plain-text files.

For each dataset, only the target numerical values specified above were included; figure labels, group names, missing values, and non-numeric annotations were removed before analysis. No transformation, normalization, rescaling, or rounding was applied after extraction. For datasets reported with fixed decimal precision, the original reported decimal format was retained. The resulting benchmark files were named RealRawData1-3 and RealErrData1-2, respectively, and were analyzed using the same feature-extraction, full-statistics, progressive subsampling, and machine-learning prediction pipeline.

2.2.4 Semi-supervised training datasets

Because confirmed fabricated research datasets are rarely available and difficult to obtain ethically, a semi-supervised simulation strategy was used to construct the training datasets. Three types of datasets were generated.

Simulated normal datasets were generated from uniform, normal, log-normal, and gamma distributions to approximate different distributional shapes and numerical ranges of continuous experimental data.

Simulated digit-preference datasets were generated by imposing predefined digit-preference probabilities on decimal digits while maintaining a continuous numerical structure. Examples included increased frequencies of digits 2, 5, and 8, or increased frequencies of two-digit combinations such as 00, 25, 50, 75, 55, and 88.

Mixed-contamination datasets were generated by combining normal simulated data with digit-preference simulated data at different contamination proportions. Given the contamination proportion $\rho$, the mixed dataset was defined as:

$$X_{mixed} = (1-\rho)X_{normal} + \rho X_{fabricated}$$

where $\rho$ was randomly varied across simulated datasets to mimic realistic scenarios in which only a subset of values is manually modified or replaced.

## 2.3 Decimal digit extraction

Let the raw numerical dataset be defined as:

$$X = \{X_1, X_2, \dots, X_n\}$$

where $n$represents the sample size. For each numerical value $X_i$, the digit at the $j$-th decimal position was extracted and denoted as:

$$D_{ij} \in \{0,1,2,\dots,9\}$$

To reduce the potential influence of real experimental effects and numerical magnitude on the leading decimal positions, the joint combination of the third and fourth decimal digits was used as the primary analysis target.

## 2.4 Single decimal-digit multinomial test

For a given decimal position $j$, the observed frequencies of digits 0–9 were denoted as:

$$O_0, O_1, \dots, O_9$$

with:

$$\sum_{k=0}^{9} O_k = n$$

Under the null hypothesis, the decimal digit follows a uniform multinomial distribution:

$$H_0: P(D_{ij} = 0) = P(D_{ij} = 1) = \cdots = P(D_{ij} = 9) = 0.1$$

The alternative hypothesis was:

$$H_1: \exists k, P(D_{ij} = k) \neq 0.1$$

The expected frequency for each digit was:

$$E_k = 0.1n$$

A Pearson chi-square goodness-of-fit test was performed:

$$\chi_j^2 = \sum_{k=0}^{9} \frac{(O_k - E_k)^2}{E_k} \text{ [30]}$$

When the sample size was sufficient and the expected frequencies met the assumptions of the test, the statistic approximately followed a chi-square distribution with 9 degrees of freedom:

$$\chi_j^2 \sim \chi^2(9)$$

To avoid overinterpretation of $P$-values, particularly in large samples, Cramér's $V$ was calculated as an effect-size measure:

$$V_j = \sqrt{\frac{\chi_j^2}{n(k-1)}}$$

where $k = 10$. Standardized residuals were also calculated to identify which digits contributed most strongly to the overall deviation:

$$R_k = \frac{O_k - E_k}{\sqrt{E_k}}$$

**2.5 Multi-decimal joint distribution test**

To detect potential combination-level preferences arising from manual fabrication or modification, a multi-decimal joint distribution test was further developed. The primary analysis focused on the combination of the third and fourth decimal digits:

$$D_{i,3:4} \in \{00, 01, 02, \dots, 99\}$$

This produced 100 possible two-digit categories. The null hypothesis was:

$$H_0: P(D_{i,3:4} = r) = 0.01, r = 00, 01, \dots, 99$$

The alternative hypothesis was that at least one two-digit combination had a probability different from 0.01.

For each combination $r$, the observed frequency was denoted as $O_r$, and the expected frequency was:

$$E_r = \frac{n}{100}$$

The chi-square statistic[30] was calculated as:

$$\chi^2_{3:4} = \sum_{r=00}^{99} \frac{(O_r - E_r)^2}{E_r}$$

with degrees of freedom:

$$df = 100 - 1 = 99$$

Cramér's $V$ was calculated as:

$$V_{3:4} = \sqrt{\frac{\chi^2_{3:4}}{n(100-1)}} \text{ [31, 32]}$$

When the sample size was small and the expected frequency per category was below 5, the conventional chi-square $P$-value was interpreted cautiously. Under such conditions, the joint distribution results were mainly interpreted using effect size, standardized entropy, KL divergence, and standardized residuals of the most abnormal digit combinations, rather than relying solely on statistical significance.

**2.6 Entropy and KL divergence**

To quantify the overall randomness of the digit distribution, Shannon entropy was calculated. Let the observed proportion of each digit or digit combination be:

$$\hat{p}_r = \frac{O_r}{n}$$

The Shannon entropy was defined as:

$$H = -\sum_r \hat{p}_r \log(\hat{p}_r) \text{ [33]}$$

where only categories with $\hat{p}_r > 0$were included. To allow comparison across different numbers of categories, normalized entropy was calculated as:

$$H^* = \frac{H}{\log(k)} \text{ [34]}$$

where $k$represents the number of possible categories. A distribution closer to the uniform distribution has $H^*$ closer to 1, whereas concentration of values in a limited number of categories leads to a lower $H^*$.

The Kullback–Leibler divergence between the observed distribution and the expected uniform distribution was also calculated:

$$D_{KL} = \sum_r \hat{p}_r \log \frac{\hat{p}_r}{p_{0r}} [35]$$

where:

$$p_{0r} = \frac{1}{k}$$

A $D_{KL}$ value closer to 0 indicates greater similarity to the theoretical uniform distribution, whereas larger values indicate stronger deviation.

**2.7 Digit-preference features**

Because manually generated numerical data may exhibit non-uniform digit-selection tendencies, several digit-preference indices were calculated. For single decimal-digit distributions, the 0/5 preference index was defined as:

$$I_{0/5} = p_0 + p_5$$

The 2/5/8 preference index was defined as:

$$I_{2/5/8} = p_2 + p_5 + p_8$$

The odd-digit preference index was defined as:

$$I_{odd} = p_1 + p_3 + p_5 + p_7 + p_9$$

The extreme-digit preference index was defined as:

$$I_{extreme} = p_0 + p_1 + p_8 + p_9$$

The middle-digit preference index was defined as:

$$I_{middle} = p_3 + p_4 + p_5 + p_6$$

For two-digit combinations, the repeated-pair index was calculated as:

$$I_{same} = p_{00} + p_{11} + p_{22} + \cdots + p_{99}$$

The neat-combination index was calculated as:

$$I_{neat} = p_{00} + p_{25} + p_{50} + p_{75}$$

The sequential-combination index was calculated as:

$$I_{seq} = p_{01} + p_{12} + p_{23} + \cdots + p_{89}$$

These indices were designed to capture interpretable digit-preference patterns that may not be fully characterized by the chi-square statistic alone.

**2.8 Progressive subsampling stability analysis**

To evaluate whether digit-distribution deviations were stable across sample sizes, progressive subsampling analysis was performed. Given a complete dataset with sample size

$N$, a sequence of increasing subsample sizes was defined:

$$n_1 < n_2 < \cdots < N$$

At each sample size $n_m$, random subsampling without replacement was repeated $B$ times. For each subsample, the following statistics were calculated:

$$\chi^2(n_m), P(n_m), V(n_m), H^*(n_m), D_{KL}(n_m)$$

For each sample-size level, the mean, median, standard deviation, and 2.5th and 97.5th percentiles of these metrics were calculated to generate progressive sampling curves.

To further quantify the direction of deviation stability, a linear regression model was fitted using $\log(n)$ as the independent variable and Cramér's $V$ as the dependent variable:

$$V(n) = \alpha + \beta \log(n) + \epsilon$$

where $\beta$ represents the trend of deviation strength as sample size increases. A decreasing $V(n)$ suggests that early deviations may be attributable mainly to small-sample random fluctuation, whereas persistently elevated $V(n)$ at larger sample sizes suggests a more stable digit-distribution deviation.

**2.9 Machine-learning feature construction**

Each dataset was treated as one independent analytical unit. For each dataset, extracted features were assembled into a machine-learning input vector:

$$\mathbf{x}_i = (x_{i1}, x_{i2}, \dots, x_{ip})$$

The features were grouped into six main classes.

The class consisted of single-decimal digit-frequency features, including the proportions of digits 0-9 at the first to fourth decimal positions.

The class consisted of multi-decimal joint-frequency features, mainly the proportions of the 100 two-digit combinations from 00 to 99 at the third and fourth decimal positions.

The class consisted of distributional-deviation features, including chi-square statistics, $P$-values, $-\log_{10} P$, Cramér's $V$, and summary measures of standardized residuals.

The class consisted of information-theoretic features, including Shannon entropy, normalized entropy, $H^*$ and KL divergence.

The class consisted of digit-preference features, including the 0/5 preference index, 2/5/8 preference index, odd-even ratio, repeated-pair index, and neat-combination index.

The class consisted of progressive subsampling features, including the mean, maximum, final value, and slope of Cramér's $V$; the mean, minimum, final value, and slope of normalized entropy; and the mean, maximum, final value, and slope of KL divergence.

Missing and infinite values were replaced before modeling. For algorithms requiring distance-based or boundary-based learning, continuous features were standardized.

**2.10 Semi-supervised machine-learning models**

Because reliable labels for real fabricated research datasets are scarce, a semi-supervised learning framework was used. Simulated normal datasets were treated as low-risk reference samples, while simulated digit-preference datasets and mixed-contamination datasets were treated as high-risk reference samples.

The following models were constructed and compared.

2.10.1 Random Forest

Random Forest was used to capture nonlinear relationships and interactions among digit-pattern features. The model consisted of multiple classification trees, each trained on a bootstrap sample, with a random subset of features considered at each split. The final class probability was obtained by averaging predictions across trees. The high-risk class probability output by Random Forest was denoted as:

$$P_{RF}$$

Variable importance was calculated to evaluate the contribution of individual digit-pattern features[36].

2.10.2 Elastic-net Logistic Regression

Elastic-net Logistic Regression was used to construct a sparse and interpretable linear risk model. The penalty term included both L1 and L2 regularization:

$$\lambda\left[\alpha \sum_j \mid \beta_j \mid + \frac{1-\alpha}{2}\sum_j \beta_j^2\right]$$

where λcontrols the overall penalty strength and $\alpha$controls the relative contribution of L1 and L2 penalties. This model was used to identify key digit-pattern features associated with high-risk classification[37].[38]

2.10.3 Support Vector Machine

Support Vector Machine was used to learn nonlinear decision boundaries between normal reference datasets and simulated irregular datasets. A radial basis function kernel was used:

$$K(x_i, x_j) = \exp\left(-\gamma \mid\mid x_i - x_j \mid\mid^2\right)$$

The model output was converted into a calibrated high-risk probability[39, 40].

2.10.4 Isolation Forest

Isolation Forest was used as an anomaly-detection model and was trained primarily on normal reference datasets to learn the boundary of normal digit structures. The model is based on the principle that anomalous samples are more easily isolated through random partitioning. Each dataset received an anomaly score, with higher scores indicating greater deviation from normal reference digit patterns[41].

2.10.5 Integrated risk model

To integrate the advantages of classification-based and anomaly-detection-based models, the final risk score was defined as:

$$Risk_{final} = \lambda Score_{ML} + (1-\lambda)Score_{IF}$$

where $Score_{ML}$ represents the high-risk probability derived from Random Forest, Elastic-net Logistic Regression, Support Vector Machine, or their weighted average; $Score_{IF}$ represents the Isolation Forest anomaly-risk score; and λ is a weighting parameter ranging from 0 to 1[42].

2.11 Model training and validation

The dataset-level feature matrix was randomly divided into training and internal testing sets at an 8:2 ratio. All models were trained using the training set and evaluated using the internal testing set. To reduce the influence of random splitting, repeated cross-validation or repeated random-split validation may be applied during model development[43].

Model performance was evaluated using the following metrics:

$$Accuracy = \frac{TP+TN}{TP+TN+FP+FN} \text{ [44]}$$

$$Sensitivity = \frac{TP}{TP+FN}$$

$$Specificity = \frac{TN}{TN+FP}$$

$$Precision = \frac{TP}{TP+FP}$$

$$F1 = \frac{2 \times Precision \times Sensitivity}{Precision + Sensitivity} \text{ [45]}$$

The area under the receiver operating characteristic curve was calculated to assess overall discrimination. Probability calibration was evaluated using the Brier score:

$$Brier\ score = \frac{1}{n}\sum_{i=1}^{n}(\hat{p}_i - y_i)^2$$

where $\hat{p}_i$ is the predicted high-risk probability and $y_i$ is the reference label. A lower Brier score indicates better calibration[46].

**2.12 Risk-grade definition**

Based on the final integrated risk score, the digit-pattern irregularity risk was classified into five levels (Table 1).

This risk grade reflects the degree of deviation from expected or reference digit patterns.

It should not be interpreted as direct evidence of data fabrication or research misconduct. Datasets with high risk scores should be further evaluated using original instrument outputs, laboratory records, data-processing workflows, rounding rules, repeated experiments, and expert review.

### 2.13 Software implementation

All analysis was performed in the R environment. Statistical testing, digit extraction, simulated data generation, progressive subsampling, feature engineering, and machine-learning modeling were implemented using custom R scripts. Random Forest models were constructed using the ranger package, Isolation Forest models were constructed using the isotree package, and visualizations were generated using the ggplot2 package. The input file was a single-column txt file, with each row representing one independent numerical value. The scripts automatically read the txt file, extracted decimal digits, calculated statistical metrics, trained machine-learning models, and exported risk scores, result tables, and visualization figures. The source code and example repository structure are available at https://github.com/YoYoCcN/FDRS-digit-randomness-screening for noncommercial academic and research use.

### 2.14 Methodological interpretation boundary

FDRS was designed as a screening tool for digit-pattern irregularities in raw numerical research data, not as a direct detector of fabrication. Non-random digit patterns may arise from manual fabrication or modification, but they may also result from benign factors such as instrument precision, fixed rounding rules, bounded measurement ranges, spreadsheet formatting, batch processing, data transformation, repeated-measurement structures, or non-malicious recording habits. Therefore, positive FDRS findings should be interpreted as signals of digit-structure irregularity requiring further verification, rather than as definitive evidence of research misconduct.

## 3 Results

### 3.1 Statistical evaluation of digit-pattern randomness in RawData and ErrData

To evaluate the statistical performance of FDRS, we first compared the digit-pattern structure of the instrument-derived RawData and the blinded manually simulated ErrData. The analysis focused on the third decimal digit and the joint distribution of the third–fourth decimal digits. RawData contained 253 enzymatic activity absorbance values, whereas ErrData contained 255 manually generated pseudo-experimental values.

For the single third-decimal-digit analysis, RawData showed no significant deviation from the theoretical uniform distribution across digits 0-9 (Table 2). The chi-square statistic was 7.9091 with 9 degrees of freedom, corresponding to $P$=0.5433. The effect size was small, with Cramér's $V$=0.0589, and the normalized entropy was close to 1 ($H^*$=0.9932), indicating a broadly uniform single-digit distribution (Figure 2A, H, I). In contrast, ErrData showed a significant deviation from the expected uniform distribution, with a chi-square statistic of 17.5882 and $P$=0.0403. ErrData also showed a higher Cramér's $V$ value (0.0875), lower normalized entropy ($H^*$=0.9851), and higher KL divergence than RawData (Figure 2A, H, I). These findings indicate that the blinded manually simulated dataset contained a detectable single-decimal-position digit imbalance.

The standardized residual analysis further supported this difference. In RawData, residuals were distributed within a relatively narrow range around zero, suggesting that no single digit strongly dominated the deviation pattern. In ErrData, several digits showed larger positive or negative residuals, indicating a more uneven contribution of specific digits to the overall chi-square statistic (Figure 2B). These results suggest that human-generated numerical values may retain subtle digit-selection preferences even when the values are intended to appear plausible and random.

**3.2 Joint third-fourth decimal digit distribution**

We next examined the joint distribution of the third and fourth decimal digits, corresponding to 100 possible two-digit categories from 00 to 99. For RawData, the expected count per category was 2.53. The joint distribution did not significantly deviate from the theoretical uniform 00-99 distribution, with $\chi^2$=84.9447, 99 degrees of freedom, and $P$=0.8419. The corresponding Cramér's $V$ was 0.0582, normalized entropy was 0.9628, and KL divergence was 0.1715 (Figure 2C, H, I; Table 4).

For ErrData, the expected count per category was 2.55. The joint third–fourth decimal digit distribution also did not reach statistical significance, with $\chi^2$=103.8235, 99 degrees of freedom, and $P$=0.3503. However, compared with RawData, ErrData showed a higher Cramér's $V$ (0.0641), lower normalized entropy $H^*$=0.9485, and higher KL divergence (0.2371) (Figure 2C, H, I; Table 3). These results indicate that ErrData had a relatively less uniform two-digit structure than RawData, although the magnitude of deviation was modest and did not reach statistical significance in the full 00-99 joint-distribution test.

The heatmap of standardized residuals for the 00-99 joint distribution showed that both datasets contained local residual variation, as expected under sparse category counts. However, ErrData displayed a visually more uneven residual structure than RawData, consistent with the higher KL divergence and lower entropy observed in the full-sample joint-distribution metrics (Figure 2C, H, I). Because the expected count per category was below 5 for both datasets, the joint-distribution chi-square $P$-values should be interpreted cautiously and in combination with effect-size, entropy, KL divergence, and residual-structure metrics.

**3.3 Full-sample multi-metric comparison**

A full-sample multi-metric summary further demonstrated that ErrData consistently showed stronger digit-pattern irregularity than RawData across several complementary metrics. In the single third-decimal-digit test, ErrData showed a larger chi-square statistic, lower $P$-value, higher Cramér's $V$, lower normalized entropy, and higher KL divergence than RawData (Figure 2H, I). In the joint third-fourth decimal digit analysis, the same overall pattern was observed, although the difference was less pronounced and did not reach statistical significance by the chi-square test (Figure 2H, I; Table 3).

These results demonstrate the value of evaluating digit randomness using multiple complementary indicators rather than relying on a single $P$-value. In particular, the combination of Cramér's $V$, normalized entropy, KL divergence, standardized residuals, and joint-combination heatmaps provided a more informative description of digit-pattern structure than the chi-square test alone (Figure 2A-C, H, I).

**3.4 Progressive subsampling analysis**

Progressive subsampling analysis was performed to determine whether digit-distribution deviations changed systematically as sample size increased. For each dataset, repeated subsampling was conducted at increasing sample sizes, and the mean chi-square statistic, median $P$-value, mean Cramér's $V$, mean normalized entropy $H^*$, mean KL divergence, and median $-\log_{10}P$ were calculated. Except for the full-sample endpoint, each subsampling level was repeated 1,000 times (Table 3).

At the smallest subsample size (n=30), RawData and ErrData showed very similar profiles, reflecting the instability of sparse 00-99 joint categories under small-sample conditions. RawData had a mean $\chi^2$of 97.6800, median $P$=0.5476, mean Cramér's $V$=0.1809, mean $H^*$=0.6993, and mean KL divergence of 1.3846. ErrData showed comparable values, with mean $\chi^2$=99.3200, median $P$=0.5476, mean Cramér's $V$=0.1825, mean $H^*$=0.6966, and mean KL divergence of 1.3972 (Table 3). These results indicate that at small subsample sizes, both datasets exhibited substantial stochastic variation due to sparse digit-combination counts.

As sample size increased, RawData showed a clear progressive movement toward the expected uniform joint distribution. The mean chi-square statistic decreased from 97.6800 at n=30 to 84.9447 at the full-sample endpoint(n=253). The median $P$-value increased from 0.5476 to 0.8419, whereas median $-\log_{10}P$ decreased from 0.2615 to 0.0747. In parallel, mean

Cramér's $V$ declined from 0.1809 to 0.0582, mean normalized entropy increased from 0.6993 to 0.9628, and mean KL divergence decreased from 1.3846 to 0.1715 (Figure 2D-G; Table 3). These changes indicate that the apparent digit-combination deviations in RawData were progressively attenuated as more observations were included, suggesting that early deviations were largely attributable to small-sample fluctuation.

ErrData showed a partially different trajectory. Mean Cramér's $V$ decreased from 0.1825 at n=30 to 0.0641 at n=255, and mean KL divergence decreased from 1.3972 to 0.2371, indicating that some sparse-sample variability was also reduced as sample size increased. However, unlike RawData, the mean chi-square statistic increased from 99.3200 to 103.8235, while the median $P$-value decreased from 0.5476 to 0.3503. Median $-\log_{10} P$ also increased from 0.2615 to 0.4556 (Figure 2D-G; Table 3). Thus, although ErrData did not reach statistical significance in the final 00-99 joint-distribution test, its evidence of deviation became relatively stronger with increasing sample size compared with RawData.

Directional-change analysis confirmed this contrast. From the smallest to the full-sample endpoint, RawData showed a decrease in $\chi^2$($\Delta\chi^2$=-12.7353), an increase in median $P$ ($\Delta P$=+0.2943), a 67.8% decrease in Cramér's $V$, an increase in $H^*$of 0.2634, and an 87.6% decrease in KL divergence (Table S1). These changes are consistent with progressive attenuation of digit-distribution deviation. In contrast, ErrData showed an increase in $\chi^2$($\Delta\chi^2$=+4.5035), a decrease in median $P$($\Delta P$=-0.1973), a 64.9% decrease in Cramér's $V$, an increase in $H^*$ of 0.2519, and an 83.0% decrease in KL divergence (Table S1). Although some dispersion metrics improved with sample size, the opposite directions of $\chi^2$, median $P$, and $-\log_{10} P$ suggest that the deviation signal in ErrData persisted relatively more strongly.

The slope analysis further supported this interpretation. In RawData, the slopes for Cramér's $V$, KL divergence, $-\log_{10} P$, and $\chi^2$ were negative, whereas the slopes for $H^*$ and median $P$ were positive. Specifically, RawData showed slopes of -0.0552 for Cramér's $V$ , +0.1203 for $H^*$, -0.5539 for KL divergence, +0.1319 for median $P$, -0.0838 for $-\log_{10} P$, and -5.9736 for $\chi^2$ (Table S2). This pattern indicates a progressive weakening of non-uniformity. By contrast, ErrData showed negative slopes for Cramér's $V$and KL divergence, but its median $P$ slope was negative (-0.0870), $-\log_{10} P$ slope was positive (+0.0855), and $\chi^2$ slope was positive (+2.1594) (Table S2). This divergence in trend direction suggests that ErrData retained a more persistent deviation tendency despite the expected reduction in sparse-sample variability.

Collectively, the progressive subsampling results demonstrate that RawData and ErrData behaved differently as sample size increased. RawData moved toward weaker evidence of deviation from the expected uniform joint distribution, whereas ErrData retained a relatively stronger and more persistent deviation signal (Figure 2D-G; Table 4; Table S1; Table S2). These findings support the use of progressive subsampling as a stability-oriented component of FDRS, allowing the framework to distinguish transient small-sample fluctuation from more persistent non-random digit-structure patterns.

**3.5 Semi-supervised machine-learning model for digit-pattern irregularity screening**

To further integrate the full-sample digit statistics and progressive subsampling features, a semi-supervised machine-learning module was incorporated into the FDRS framework. The model used dataset-level features derived from single-digit distributions, joint third–fourth decimal digit distributions, information-theoretic metrics, digit-preference indices, and progressive subsampling trajectories. These features were extracted from simulated normal datasets, digit-preference irregular datasets, and mixed-contamination datasets, and were then used to train Random Forest, Elastic-net Logistic Regression, radial-basis-function Support Vector Machine, Isolation Forest, and an integrated ensemble model.

The ROC analysis showed that most supervised classifiers achieved strong discriminatory performance in the internal validation set (Figure 3A). Elastic-net Logistic Regression achieved the highest AUC, with an AUC of 0.98395, followed by Random Forest with an AUC of 0.98090, the ensemble model with an AUC of 0.97895, and SVM radial with an AUC of 0.97540 (Table 4). Isolation Forest showed weaker but still meaningful discrimination, with an AUC of 0.89220, indicating that anomaly-boundary learning alone was less powerful than supervised classifiers but remained informative as an auxiliary risk component.

Calibration analysis further showed that Elastic-net Logistic Regression had the best probability calibration, with the lowest Brier score of 0.048439 (Figure 3B; Table 4). Random Forest and SVM radial also showed favorable calibration, with Brier scores of 0.055680 and 0.056181, respectively. The ensemble model yielded a Brier score of 0.065783. In contrast, Isolation Forest had the highest Brier score of 0.206379, suggesting less reliable probability calibration when used as a standalone anomaly detector.

Overall classification performance was consistent with the ROC and calibration results. Random Forest achieved the highest overall accuracy (0.926667), specificity (0.960), precision (0.978495), F1 score (0.943005), and balanced accuracy (0.935) (Table 4). Elastic-net Logistic Regression showed the highest sensitivity (0.915), indicating slightly stronger detection of irregular digit-pattern datasets. SVM radial also performed well, with an accuracy of 0.916667, specificity of 0.950, precision of 0.972973, and AUC of 0.97540. The ensemble model achieved an accuracy of 0.913333, sensitivity of 0.910, specificity of 0.920, and AUC of 0.97895. The confusion matrices confirmed that the supervised models and the ensemble model correctly classified most normal and irregular reference datasets, whereas Isolation Forest produced relatively more classification errors (Figure 3E). These results indicate that the digit-pattern features extracted by FDRS can be effectively learned by machine-learning models, particularly by Elastic-net Logistic Regression and Random Forest.

### 3.6 Feature-space visualization and feature-importance analysis

PCA of the digit-pattern feature space showed partial separation among simulated normal datasets, digit-preference irregular datasets, and mixed-contamination datasets (Figure 3C). Although the feature space showed some overlap, the target datasets occupied distinct positions relative to the training distributions. RawData was positioned closer to the simulated normal reference region, whereas ErrData was shifted toward the mixed or irregular feature space. This spatial pattern was consistent with the downstream model-derived risk scores.

Feature-importance analysis indicated that the model relied on multiple classes of digit-pattern features rather than a single isolated metric (Figure 3D). The top-ranked features included variables derived from single-digit distributions, joint digit-combination statistics, information-theoretic measures, and progressive subsampling descriptors. This supports the rationale of FDRS as a multi-feature screening framework in which full-sample statistics and progressive stability metrics jointly contribute to risk discrimination. The full-sample statistical summary and progressive subsampling plots also showed that RawData and ErrData differed across several complementary metrics, including chi-square statistics, $P$-value trajectories, Cramér's $V$, normalized entropy, and KL divergence (Figure 3I, K).

### 3.7 Risk prediction for RawData and ErrData

The trained semi-supervised machine-learning models were then applied to the two target datasets: the instrument-derived RawData and the blinded manually simulated ErrData. The two datasets showed clearly different risk profiles across most models (Figure 3F; Table 5).

For RawData (n=253), all model-derived risk scores were low. Random Forest assigned a risk score of 0.170373, Elastic-net Logistic Regression assigned 0.149674, SVM radial assigned 0.054441, and Isolation Forest assigned 0.124020. The final ensemble risk score was 0.124627, classifying RawData as Grade 0: no apparent irregularity (Table 5). This low-risk classification was consistent with the statistical analysis showing that RawData did not significantly deviate from the expected digit distribution and that its progressive subsampling trajectory moved toward weaker evidence of non-uniformity.

In contrast, ErrData (n=255) received substantially higher risk scores from the supervised classifiers. Random Forest assigned a risk score of 0.787406, Elastic-net Logistic Regression assigned 0.950056, and SVM radial assigned 0.923895. Isolation Forest produced a lower but still elevated score of 0.301684, suggesting that the anomaly-boundary model detected weaker separation than the supervised classifiers. After integration, the final ensemble risk score for ErrData was 0.740760, corresponding to Grade 3: high irregularity (Figure 3F; Table 5). These findings indicate that the manually simulated irregular dataset contained digit-pattern features that were strongly recognized by the trained FDRS machine-learning module.

The Isolation Forest anomaly-score distributions provided an additional perspective on target-dataset behavior (Figure 3G, J). RawData was located within the low-anomaly region close to the simulated normal reference distribution. ErrData shifted toward a higher anomaly-risk region, although the separation was less pronounced than that observed for the supervised models. This explains why Isolation Forest contributed a moderate risk score for ErrData, while Random Forest, Elastic-net Logistic Regression, and SVM radial assigned much higher risk scores.

Collectively, the machine-learning results demonstrate that FDRS can integrate full-sample digit statistics, progressive subsampling features, and digit-preference indicators into a quantitative risk score. RawData was consistently assigned a low-risk profile and classified as Grade 0, whereas ErrData was assigned a high-risk profile and classified as Grade 3. These results support the feasibility of FDRS as an auxiliary screening tool for identifying non-random digit-pattern irregularities in raw numerical research data, while the resulting risk grade should be interpreted as a screening signal rather than direct evidence of data fabrication.

### 3.8 External real-world benchmark validation

To further evaluate the applicability of FDRS beyond the proof-of-concept RawData and ErrData datasets, the trained model was applied to five independent real-world benchmark datasets curated from published source data. These datasets were not used during model training, parameter tuning, or internal validation. The benchmark set included three RealRawData datasets, defined as published numerical datasets without identified public post-publication concerns for the analyzed records, and two RealErrData datasets, selected from articles or figure panels with publicly documented data-integrity concerns or institutional/journal-level handling records (Table 6, Figure S2, S3).

The three RealRawData datasets showed low-to-mild risk profiles. RealRawData1 contained 221 numerical values and received low risk scores from the supervised classifiers, including 0.197956 from Random Forest, 0.017005 from Elastic-net Logistic Regression, and 0.013691 from SVM radial. Although Isolation Forest assigned a slightly higher score of 0.221924, the final ensemble risk score was 0.112644, corresponding to Grade 0: no apparent irregularity. RealRawData2 contained 124 values and showed a mild-risk profile, with model scores of 0.470774 from Random Forest, 0.049999 from Elastic-net Logistic Regression, 0.133819 from SVM radial, and 0.373728 from Isolation Forest. Its final ensemble risk score was 0.257080, corresponding to Grade 1: mild irregularity. RealRawData3 contained 266 values and also received a Grade 1 classification, with an ensemble risk score of 0.270803.

In contrast, the two RealErrData datasets showed higher risk grades. RealErrData1 contained 341 numerical values and received intermediate risk scores from Random Forest and Elastic-net Logistic Regression, with values of 0.430195 and 0.428039, respectively. SVM radial assigned a lower score of 0.215075, whereas Isolation Forest assigned a higher anomaly-risk score of 0.717983. The integrated ensemble risk score was 0.447823, corresponding to Grade 2: moderate irregularity. RealErrData2 contained 470 two-decimal-place numerical values and showed the highest overall risk among the benchmark datasets. Random Forest assigned a score of 0.743920, Elastic-net Logistic Regression assigned 0.999978, and Isolation Forest assigned 1.000000, whereas SVM radial assigned a lower score of 0.215075. The final ensemble risk score was 0.739743, corresponding to Grade 3: high irregularity.

Overall, all three RealRawData datasets were classified as Grade 0 or Grade 1, whereas both RealErrData datasets were classified as Grade 2 or Grade 3. This pattern suggests that FDRS can provide a graded external benchmark signal that is broadly concordant with the predefined real-world dataset categories. However, the model did not produce a binary separation between all benchmark classes. The mild-risk scores observed for RealRawData2 and RealRawData3 indicate that low-level digit-structure irregularity may also arise in published datasets without known public concerns, potentially reflecting reporting precision, measurement workflow, sample-size structure, rounding, or data-format constraints. Conversely, the moderate rather than extreme score of RealErrData1 suggests that publicly questioned datasets may contain heterogeneous numerical irregularity patterns that are not always dominated by the digit-preference features learned during simulation. Therefore, the real-world benchmark results support the use of FDRS as a prioritization tool for further review rather than as a standalone adjudicative test.

## 4 Discussion

This study developed and preliminarily validated FDRS, a statistical–machine learning framework for screening non-random digit patterns in raw numerical research data. The framework integrates single-decimal-digit testing, joint third–fourth decimal digit distribution analysis, information-theoretic metrics, progressive subsampling stability assessment, and semi-supervised machine-learning risk scoring. The central aim of FDRS is not to determine whether a dataset has been fabricated, but to identify digit-structure irregularities that may warrant further verification. The current results support this position. The single-digit module detected a significant third decimal-position deviation in the blinded manually simulated ErrData, whereas the instrument-derived RawData showed no such deviation. In the joint third–fourth decimal digit analysis, neither dataset reached statistical significance by the 00-99 chi-square test, but ErrData showed relatively higher Cramér's V, lower normalized entropy, and higher KL divergence than RawData. These findings indicate that the manually simulated dataset contained stronger non-random digit-structure features, although the magnitude of the joint-distribution deviation remained modest.

The primary research value of the FDRS lies in the fact that it extends the scope of research integrity reviews from traditional images, text, and summary statistics to raw numerical data. Previous research on research integrity has shown that data fabrication and data manipulation are not entirely uncommon. Fanelli's meta-analysis revealed that approximately 1.97% of researchers admitted to having fabricated, falsified, or altered data or results at least once; simultaneously, a higher proportion of researchers reported having observed similar behavior in colleagues[3]. In the field of clinical research, efforts have been made to use statistical methods to identify suspicious data patterns. For example, a study by Al-Marzouki et al. published in the BMJ examined the application of statistical methods in detecting data fabrication in clinical trials[17], while Carlisle conducted a large-scale statistical screening of non-random sampling and potential data anomalies in 5,087 randomized controlled trials[18]. These studies demonstrate that statistical anomalies cannot be directly equated with conclusions of fabrication, but they can serve as important clues prompting further verification[17, 18].

Unlike traditional leading-digit analysis based on Benford's Law, FDRS is better suited for raw laboratory data with a fixed decimal precision. Benford-type methods are generally more applicable to naturally generated numerical data spanning multiple orders of magnitude, whereas laboratory absorbance, fluorescence values, Ct values, enzyme activity readings, or grayscale quantification values typically have a narrower numerical range and are often reported with a fixed number of decimal places[11]. For such data, the distribution of the leading digit may not have a suitable theoretical reference, whereas the trailing decimal places are more likely to reflect measurement noise, recording processes, and random fluctuations under specific instrument precision and data generation mechanisms. FDRS capitalizes on this by focusing the analysis on the third and fourth decimal places and further constructing a joint 00–99 combination distribution[13]. Previous research by Mosimann et al. has also shown that even when people subjectively attempt to generate random numbers, manually generated numbers often deviate from a truly uniform distribution[14], providing empirical support for the hypothesis that "manually generated numbers may leave statistical traces."

A major contribution of FDRS is its use of progressive subsampling to distinguish transient small-sample fluctuation from more persistent digit-pattern irregularity. At small subsample sizes, both RawData and ErrData showed substantial variation in the 00-99 joint distribution, which is expected because the number of categories is large and the expected count per category is sparse. However, the two datasets diverged as sample size increased. RawData progressively moved toward weaker evidence of deviation, as reflected by decreasing chi-square, decreasing Cramér's $V$, increasing median $P$, increasing normalized entropy, and decreasing KL divergence. In contrast, ErrData showed decreasing Cramér's $V$ and KL divergence but increasing chi-square, decreasing median $P$, and increasing $-\log_{10}P$. The directional-change and slope analyses further confirmed this contrast: RawData showed an attenuation pattern, whereas ErrData retained a relatively more persistent deviation signal. This result is important because it demonstrates that FDRS does not rely on a single full-sample $P$-value, but instead evaluates whether irregularity signals remain stable or dissipate as more observations are included.

The progressive subsampling results also illustrate why digit-pattern screening requires multiple complementary metrics. In the joint 00-99 test, both RawData and ErrData had expected counts below 5 per category, making the conventional chi-square $P$-value less reliable as a standalone decision criterion. Under this condition, effect-size and information-theoretic metrics become particularly useful. Cramér's $V$ provides a scale-adjusted measure of deviation strength, normalized entropy reflects the degree of distributional dispersion, and KL divergence quantifies departure from the expected uniform distribution. The fact that ErrData consistently showed lower entropy and higher KL divergence than RawData supports the interpretation that its digit-combination structure was less uniform, even when the final joint-distribution $P$-value did not reach statistical significance. Therefore, FDRS should be understood as a multi-metric screening framework rather than a binary hypothesis-testing procedure.

The semi-supervised machine-learning module further improved the ability of FDRS to integrate complex digit-pattern features. In the internal validation set, most supervised classifiers achieved strong discrimination between simulated normal and irregular datasets. Elastic-net Logistic Regression achieved the highest AUC of 0.98395 and the lowest Brier score of 0.048439, indicating both excellent discrimination and favorable probability calibration. Random Forest showed the highest overall accuracy, specificity, precision, F1 score, and balanced accuracy, suggesting strong and balanced classification performance. SVM radial also performed well, with an AUC of 0.97540 and high specificity and precision. The ensemble model achieved an AUC of 0.97895, supporting the feasibility of integrating multiple classifiers into a unified risk-score framework. By contrast, Isolation Forest showed weaker standalone performance, with an AUC of 0.89220 and a Brier score of 0.206379, indicating that unsupervised anomaly-boundary detection alone was less powerful and less calibrated than supervised classifiers in the current feature space.

The updated target-dataset predictions provide a more nuanced interpretation than the preliminary version of the model. RawData received consistently low risk scores from all models: 0.170373 by Random Forest, 0.149674 by Elastic-net Logistic Regression, 0.054441 by SVM radial, and 0.124020 by Isolation Forest. The final ensemble risk score was 0.124627, classifying RawData as Grade 0, indicating no apparent digit-pattern irregularity. This low-risk classification is consistent with the instrument-derived origin of RawData and with its statistical profile, in which single-digit testing did not show significant deviation and progressive subsampling showed attenuation of digit-distribution irregularity. In contrast, ErrData received high risk scores from the supervised classifiers, including 0.787406 by Random Forest, 0.950056 by Elastic-net Logistic Regression, and 0.923895 by SVM radial. The final ensemble risk score was 0.740760, classifying ErrData as Grade 3, indicating high irregularity. This result demonstrates that the manually simulated dataset contained digit-pattern features that were strongly recognized by the supervised FDRS models.

The PCA visualization further supports the interpretation that FDRS captures structured feature-space differences rather than isolated statistical artifacts. The simulated normal, mixed-contamination, and digit-preference irregular datasets showed partially separated but overlapping regions, indicating that digit-pattern irregularity exists on a continuum rather than as a perfectly separable binary state. RawData was located closer to the normal reference space, whereas ErrData shifted toward the mixed or irregular feature region. This pattern is consistent with the model-derived risk scores and reinforces the idea that FDRS should be used to assign relative screening priority rather than to make categorical claims of fabrication.

The feature-importance analysis also highlights the interpretability of the FDRS framework. The top-ranked features were not limited to a single statistical test; rather, they included features derived from single-digit distributions, joint digit-combination statistics, information-theoretic metrics, digit-preference indices, and progressive subsampling descriptors. This supports the conceptual design of FDRS as an interpretable feature-engineering framework. Unlike black-box models that directly classify raw numerical columns, FDRS first converts the dataset into statistically meaningful descriptors, such as digit frequencies, Cramér's $V$, entropy, KL divergence, residual structures, preference indices, and progressive slopes. As a result, high-risk outputs can be traced back to specific digit-pattern behaviors, which is essential in research-integrity applications where algorithmic decisions must remain transparent and auditable.

From the perspective of research integrity, FDRS extends existing statistical-forensic approaches from published summary statistics to raw numerical datasets. Tools such as GRIM and stat check focus on mathematical consistency in reported results[7, 8], whereas FDRS operates directly on raw numerical columns. This makes it potentially useful for datasets such as absorbance readings, qPCR Ct values, fluorescence intensities, cell-counting outputs, ELISA measurements, image-derived quantitative values, enzyme-activity readouts, and high-throughput screening data. Compared with Benford-type leading-digit analysis, FDRS is better suited to fixed-decimal laboratory measurements, where values usually occupy a limited numerical range and leading-digit distributions may not have a meaningful theoretical reference[11]. In such contexts, later decimal digits and multi-digit combinations may provide more relevant information about fine-scale numerical structure.

The real-world benchmark analysis provides an important external context for interpreting FDRS. Unlike the controlled proof-of-concept comparison between instrument-derived RawData and manually simulated ErrData, the RealRawData and RealErrData datasets were heterogeneous in data type, numerical precision, measurement source, and publication context. Despite this heterogeneity, the benchmark results showed an ordinal risk pattern: all three RealRawData datasets were classified as Grade 0 or Grade 1, whereas the two RealErrData datasets were classified as Grade 2 or Grade 3. This finding suggests that the FDRS risk score captures digit-structure features that are at least partially concordant with real-world data-integrity concern status.

However, these results should be interpreted cautiously. The RealRawData label does not prove that a dataset is free of error or manipulation; it only indicates that no public post-publication concern or formal handling record was identified for the specific analyzed numerical records at the time of analysis. Similarly, the RealErrData label does not mean that every numerical value in the selected panels was fabricated; it indicates that the corresponding articles or figure panels were associated with documented post-publication concerns, corrections, or institutional/journal-level handling records. Therefore, the benchmark labels are imperfect external reference categories rather than definitive ground-truth labels. The Grade 1 classifications of RealRawData2 and RealRawData3 further emphasize that mild digit-pattern irregularity can occur in apparently unchallenged published numerical datasets, possibly due to benign factors such as fixed reporting precision, bounded measurement scales, rounding conventions, or data-processing workflows.

The model-specific scores also revealed heterogeneous detection behavior across benchmark datasets. For example, RealErrData2 received very high risk scores from Random Forest, Elastic-net Logistic Regression, and Isolation Forest, resulting in a Grade 3 ensemble classification. In contrast, RealErrData1 showed a more moderate ensemble score, with stronger contribution from Isolation Forest than from SVM radial. This suggests that different irregularity mechanisms may be captured by different model components. Supervised classifiers may be more sensitive to digit-preference patterns resembling the simulated irregular training data, whereas Isolation Forest may detect broader deviations from the normal-reference feature space. Thus, the ensemble structure is useful because it integrates complementary signals rather than relying on a single model output.

Several limitations should be acknowledged. The training framework relied on simulated normal, digit-preference, and mixed-contamination datasets. Although these simulations cover several plausible forms of digit irregularity, they cannot fully represent the diversity of real-world data manipulation or data-quality problems, such as local copy-and-paste duplication, selective deletion, variance compression, excessive smoothing, significance-driven adjustment, over-randomization, or hybrid manual-instrumental workflows. The proof-of-concept validation used only one instrument-derived absorbance dataset and one blinded manually simulated dataset. This design is useful for initial feasibility testing but is insufficient to establish generalizability across experimental platforms and research fields. Over, FDRS currently produces dataset-level risk scores and does not identify which specific values, if any, are problematic. Although the present study added five real-world benchmark datasets, the external validation remains preliminary. The benchmark datasets were manually curated from published source data rather than from original instrument files or laboratory information systems. The RealRawData and RealErrData categories were based on public concern status and available handling records, not on complete independent forensic adjudication of every numerical record. Therefore, these labels should be regarded as pragmatic benchmark categories rather than definitive ground truth. In addition, the number of benchmark datasets was small, the data types were heterogeneous, and the tested decimal positions differed according to reporting precision. Larger benchmark collections across different experimental platforms are needed to calibrate risk thresholds, estimate false-positive and false-negative rates, and determine whether dataset-specific digit-position selection improves generalizability.

Future work should focus on expanding both the reference data and the irregularity models. Larger multi-source databases of authentic raw numerical data should be built from diverse experimental platforms, including microplate-reader absorbance, qPCR, ELISA, flow cytometry, microscopy quantification, dose–response assays, and high-throughput screening. The simulation framework should also be expanded to include weak digit preference, strong digit preference, local contamination, repeated-value insertion, copy-paste blocks, excessive rounding, variance compression, over-smoothing, and overly uniform artificial randomness. In addition, future versions of FDRS should incorporate external validation cohorts, SHAP-based interpretability, Bayesian risk modeling, likelihood-ratio frameworks, and calibration strategies tailored to different data-generating mechanisms.

In summary, our results demonstrate that FDRS can integrate full-sample digit statistics, progressive subsampling behavior, and machine-learning risk modeling into an interpretable screening framework for raw numerical research data. In the proof-of-concept comparison, instrument-derived RawData was classified as Grade 0, whereas manually simulated ErrData was classified as Grade 3. In the external real-world benchmark analysis, all RealRawData datasets were classified as Grade 0 or Grade 1, whereas both RealErrData datasets were classified as Grade 2 or Grade 3. These findings support the feasibility of FDRS as an auxiliary tool for identifying non-random digit-pattern irregularities and prioritizing datasets for further review. However, FDRS should not be used as a standalone detector of fabrication or misconduct; its outputs should always be interpreted in combination with original instrument files, laboratory records, data-processing workflows, repeated experiments, public correction records, and expert review.

## CRediT authorship contribution statement

**Zhuohua Cao:** Conceptualization, Data curation, Formal analysis, Funding Acquisition, Methodology, Investigation, Project Administration, Resources, Software, Supervision, Validation, Visualization, Writing – original draft preparation.

## Funding

This work was supported by personal funds from Zhuohua Cao. No specific grant from any funding agency in the public, commercial, or not-for-profit sectors was received.

## Declaration of Competing Interests

The author declares the following potential competing interest: Zhuohua Cao is an applicant/inventor on pending patent applications related to the digit-randomness screening methods and data-fabrication risk assessment framework described in this manuscript. The author declares that the patent application process did not influence the design, conduct, analysis, interpretation, or reporting of this study.

## Data availability

The RawData and ErrData datasets used in this study, the curated real-world benchmark datasets derived from published source data, and the R scripts for digit extraction, statistical testing, progressive subsampling, simulated dataset generation, and semi-supervised machine-learning analysis are provided as Supplementary Materials. The source code is also available at GitHub: https://github.com/YoYoCcN/FDRS-digit-randomness-screening. Additional information is available from the corresponding author upon reasonable request.

**Figure 1. Overall workflow of the Fabrication-risk Digit Randomness Screening model.**

The FDRS framework was designed to screen non-random digit-pattern irregularities in raw numerical research data. Raw numerical values arranged in a single-column input file were first processed for decimal digit extraction. Single-decimal-digit distributions and multi-decimal joint digit combinations were then evaluated using multinomial goodness-of-fit testing, chi-square statistics, Cramér's $V$, standardized residuals, entropy, normalized entropy, and Kullback–Leibler divergence. Progressive subsampling analysis was further performed to assess whether digit-distribution deviations attenuated or persisted as sample size increased. Dataset-level features derived from digit frequencies, digit-preference indices, information-theoretic metrics, residual patterns, and progressive-curve descriptors were subsequently used to train semi-supervised machine-learning models, including Random Forest, Elastic-net Logistic Regression, Support Vector Machine, Isolation Forest, and an integrated ensemble model. The final output was an integrated digit-pattern irregularity risk score and corresponding risk grade. FDRS is intended as an auxiliary screening and prioritization tool rather than standalone evidence of data fabrication or research misconduct.

**Figure 2. Statistical and progressive evaluation of digit-pattern randomness in RawData and ErrData.**

**(A)** Observed distribution of the single third-decimal digit in the instrument-derived RawData and the blinded manually simulated ErrData. Digits 0–9 were tested against the expected uniform distribution. **(B)** Standardized residuals from the single third-decimal-digit multinomial test, showing the contribution of each digit to the overall deviation. **(C)** Heatmap of standardized residuals for the joint third–fourth decimal digit distribution across 100 possible combinations from 00 to 99. **(D–G)** Progressive subsampling trajectories for joint third–fourth decimal digit statistics across increasing sample sizes, including chi-square statistics, median *P* values, Cramér's V, normalized entropy, KL divergence, and/or -$\log_{10}P$, as displayed in the corresponding panels. Shaded regions indicate the empirical variation across repeated subsampling iterations, whereas the full-sample endpoint was calculated once using all available values. **(H)** Full-sample multi-metric comparison of RawData and ErrData for single-digit and joint-digit analyses. **(I)** Summary visualization of information-theoretic and effect-size metrics, including Cramér's V, normalized entropy, and KL divergence.

**Figure 3. Semi-supervised machine-learning modeling and digit-pattern irregularity risk prediction.**

**(A)** Receiver operating characteristic curves of the machine-learning models in the internal validation set. AUC values are shown for Random Forest, Elastic-net Logistic Regression, SVM radial, Isolation Forest, and the integrated ensemble model. **(B)** Calibration performance of the machine-learning models, evaluated using predicted risk probabilities and Brier scores. **(C)** Principal component analysis of the digit-pattern feature space. Simulated normal, digit-preference irregular, and mixed-contamination datasets are shown as reference training distributions, with target datasets projected into the same feature space. **(D)** Feature-importance ranking showing the digit-pattern features that contributed most strongly to model discrimination. **(E)** Confusion matrices of different machine-learning models in the internal validation set. **(F)** Model-specific and ensemble risk scores for RawData and ErrData. **(G, J)** Isolation Forest anomaly-score distributions showing the relative anomaly position of target datasets compared with reference training patterns. **(H-K)** Additional model-summary and feature-space visualizations, including full-sample statistics, progressive subsampling descriptors, and integrated risk-score distributions.

**Figure S1. Feature-space projection and Isolation Forest anomaly-risk assessment of real-world benchmark datasets.**

**(A-O)** Feature-space visualization and Isolation Forest-based anomaly assessment for five independent real-world benchmark datasets. Panels **A-C** show RealRawData1, panels **D-F** show RealRawData2, panels **G-I** show RealRawData3, panels J-L show RealErrData1, and panels **M-O** show RealErrData2. For each dataset, the left panel shows principal component analysis of the digit-pattern feature space, in which simulated normal, simulated digit-preference irregular, and mixed-contamination training datasets are displayed together with the target dataset. The middle panel shows the Isolation Forest anomaly-risk density distribution of the training datasets, with the target dataset marked by a vertical line. The right panel shows the Isolation Forest anomaly-risk distribution across training categories using violin/box plots, with the corresponding target dataset overlaid as an individual point.

**Figure S2. Model-derived risk scores, full-sample digit statistics, and progressive subsampling profiles of real-world benchmark datasets.**

**(A-O)** Integrated FDRS evaluation of five independent real-world benchmark datasets. Panels **A-C** show RealRawData1, panels **D-F** show RealRawData2, panels **G-I** show RealRawData3, panels **J-L** show RealErrData1, and panels **M-O** show RealErrData2. For each dataset, the left panel shows model-specific and ensemble risk scores generated by Elastic-net Logistic Regression, Random Forest, SVM radial, Isolation Forest, and the integrated ensemble model. Horizontal dashed lines indicate predefined risk-grade thresholds. The middle panel shows the full-sample statistical summary of the target dataset, including single-decimal-digit and joint two-decimal-digit analyses. The displayed metrics include chi-square statistics, *P* values, Cramér's *V*, normalized entropy $H^*$, KL divergence, and digit-preference indices. The right panel shows progressive subsampling trajectories across increasing sample sizes, including $-\log_{10}P$, chi-square statistic, Cramér's *V*, KL divergence, median P value, and normalized entropy $H^*$.

**Table 1. Definition of integrated digit-pattern irregularity risk grades.**

The final FDRS ensemble risk score was mapped to five predefined risk grades. Scores below 0.20 were classified as Grade 0, indicating no apparent digit-pattern irregularity; scores from 0.20 to 0.40 were classified as Grade 1, indicating mild irregularity; scores from 0.40 to 0.60 were classified as Grade 2, indicating moderate irregularity and the need for further review; scores from 0.60 to 0.80 were classified as Grade 3, indicating high irregularity and the need to check raw records; and scores ≥0.80 were classified as Grade 4, indicating very high irregularity and the need for detailed verification.

**Table 2. Single third-decimal-digit distribution analysis in RawData and ErrData.**

The third decimal digit of each numerical value was extracted and tested against the expected uniform distribution across digits 0-9 using a chi-square goodness-of-fit test. RawData contained 253 instrument-derived absorbance values and showed no significant deviation from the expected uniform distribution. ErrData contained 255 blinded manually simulated pseudo-experimental values and showed a significant deviation. Cramér's V was calculated as an effect-size measure, and normalized entropy was used to quantify the dispersion of the digit distribution.

**Table 3. Progressive subsampling analysis of the joint third–fourth decimal digit distribution.** Progressive subsampling was performed to evaluate whether joint digit-distribution deviations attenuated or persisted as sample size increased. For each subsample size, repeated random subsampling without replacement was performed except at the full-sample endpoint, where all available values were analyzed once. The table summarizes mean chi-square statistics, median *P* values, mean Cramér's V, mean normalized entropy, mean KL divergence, and median $-\log_{10}P$. RawData showed progressive attenuation of digit-combination deviation as sample size.

**Table 4. Internal validation performance of the semi-supervised machine-learning models.** Performance metrics are shown for Random Forest, Elastic-net Logistic Regression, SVM radial, Isolation Forest, and the integrated ensemble model. The reported threshold corresponds to the model-specific decision cutoff used to classify datasets into low-risk or high-risk reference categories. Accuracy, sensitivity, specificity, precision, negative predictive value, F1 score, balanced accuracy, AUC, and Brier score were calculated in the internal validation set.

**Table 5. Model-derived risk scores and integrated risk grades for RawData and ErrData.** The trained FDRS models were applied to the instrument-derived RawData and the blinded manually simulated ErrData. Model-specific risk scores were generated using Random Forest, Elastic-net Logistic Regression, SVM radial, and Isolation Forest. The final ensemble risk score was calculated by integrating supervised-model risk and Isolation Forest anomaly-risk information.

**Table 6. FDRS risk prediction in independent real-world benchmark datasets.** Five independent real-world numerical datasets curated from published source data were analyzed using the trained FDRS pipeline. Model-specific risk scores from Random Forest, Elastic-net Logistic Regression, SVM radial, and Isolation Forest were integrated into a final ensemble risk score.

**Table S1. Directional changes in progressive subsampling metrics from the smallest subsample to the full-sample endpoint.** This table summarizes changes in major joint digit-distribution metrics between the smallest subsampling level and the full-sample endpoint for RawData and ErrData. $\Delta\chi^2$, $\Delta$median *P*, $\Delta$Cramér's V, $\Delta H^*$, and $\Delta$KL were calculated to characterize whether digit-distribution deviation attenuated or persisted as sample size increased.

**Table S2. Linear trend slopes of progressive subsampling metrics.**

Linear slopes were calculated for major progressive subsampling metrics across increasing sample sizes. Negative slopes for Cramér's *V*, KL divergence, $-\log_{10}P$, and $\chi^2$, together with positive slopes for $H^*$ and median *P*, indicate progressive weakening of digit-distribution deviation.

Figure 1

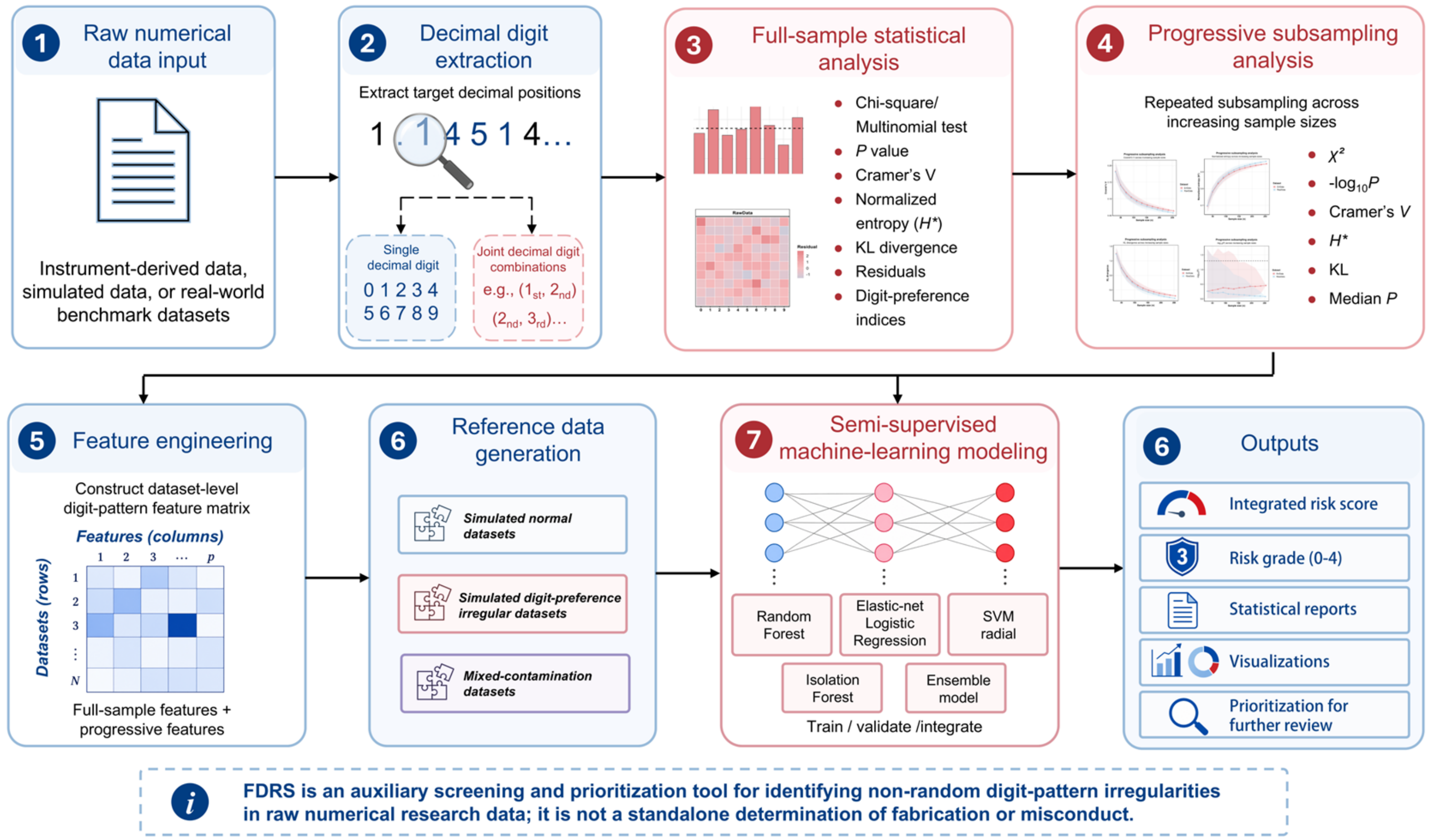

Overall workflow of the Fabrication-risk Digit Randomness Screening (FDRS) framework
1 Raw numerical data input
Instrument-derived data, simulated data, or real-world benchmark datasets
2 Decimal digit extraction
Extract target decimal positions
1 . 1 4 5 1 4…
Single decimal digit
0 1 2 3 4
5 6 7 8 9
Joint decimal digit combinations
e.g., (1st, 2nd)
(2nd, 3rd)…
3 Full-sample statistical analysis
Chi-square/ Multinomial test
P value
Cramer's V
Normalized entropy (H*)
KL divergence
Residuals
Digit-preference indices
4 Progressive subsampling analysis
Repeated subsampling across increasing sample sizes
X²
-log10P
Cramer's V
H*
KL
Median P
5 Feature engineering
Construct dataset-level digit-pattern feature matrix
Features (columns)
Datasets (rows)
Full-sample features + progressive features
6 Reference data generation
Simulated normal datasets
Simulated digit-preference irregular datasets
Mixed-contamination datasets
7 Semi-supervised machine-learning modeling
Random Forest
Elastic-net Logistic Regression
SVM radial
Isolation Forest
Ensemble model
Train / validate /integrate
6 Outputs
Integrated risk score
Risk grade (0-4)
Statistical reports
Visualizations
Prioritization for further review
FDRS is an auxiliary screening and prioritization tool for identifying non-random digit-pattern irregularities in raw numerical research data; it is not a standalone determination of fabrication or misconduct.

Figure 2

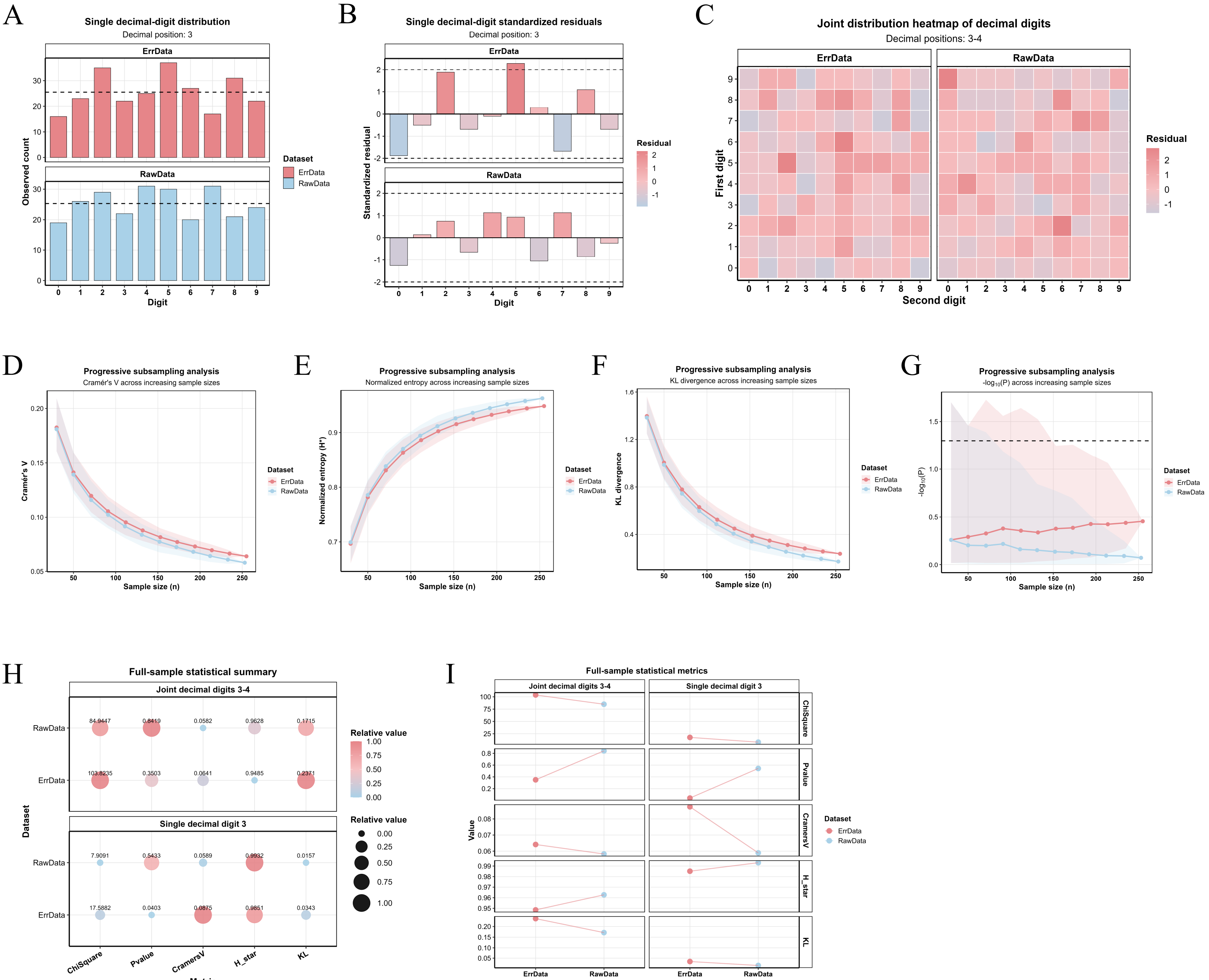

A
Single decimal-digit distribution
Decimal position: 3
ErrData
RawData
Observed count
Digit
Dataset
ErrData
RawData
B
Single decimal-digit standardized residuals
Decimal position: 3
ErrData
RawData
Standardized residual
Digit
Residual
C
Joint distribution heatmap of decimal digits
Decimal positions: 3-4
ErrData
RawData
First digit
Second digit
Residual
D
Progressive subsampling analysis
Cramér's V across increasing sample sizes
Cramér's V
Sample size (n)
Dataset
ErrData
RawData
E
Progressive subsampling analysis
Normalized entropy across increasing sample sizes
Normalized entropy (H*)
Sample size (n)
Dataset
ErrData
RawData
F
Progressive subsampling analysis
KL divergence across increasing sample sizes
KL divergence
Sample size (n)
Dataset
ErrData
RawData
G
Progressive subsampling analysis
-log10(P) across increasing sample sizes
-log10(P)
Sample size (n)
Dataset
ErrData
RawData
H
Full-sample statistical summary
Joint decimal digits 3-4
RawData 84.9447 0.8419 0.0582 0.9628 0.1715
ErrData 103.8235 0.3503 0.0641 0.9485 0.2371
Single decimal digit 3
RawData 7.9091 0.5433 0.0589 0.9932 0.0157
ErrData 17.5882 0.0403 0.0875 0.9851 0.0343
ChiSquare
Pvalue
CramersV
H_star
KL
Metric
Dataset
Relative value
I
Full-sample statistical metrics
Joint decimal digits 3-4
Single decimal digit 3
ChiSquare
Pvalue
CramersV
H_star
KL
Value
ErrData
RawData
Dataset
ErrData
RawData

# Figure 3

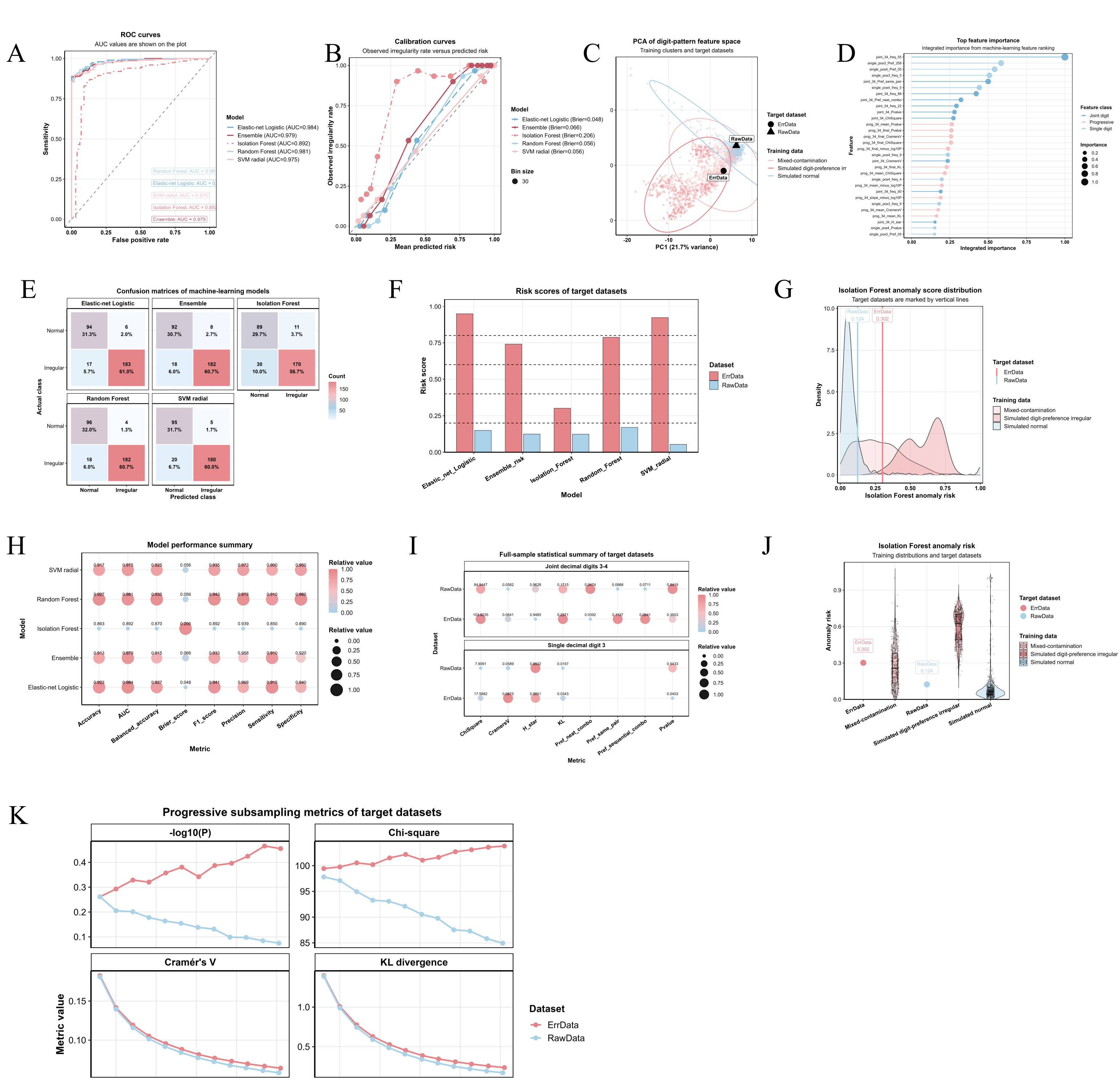
A
ROC curves
AUC values are shown on the plot
Sensitivity
False positive rate
Model
Elastic-net Logistic (AUC=0.984)
Ensemble (AUC=0.979)
Isolation Forest (AUC=0.892)
Random Forest (AUC=0.981)
SVM radial (AUC=0.975)
B
Calibration curves
Observed irregularity rate versus predicted risk
Observed irregularity rate
Mean predicted risk
Model
Elastic-net Logistic (Brier=0.048)
Ensemble (Brier=0.066)
Isolation Forest (Brier=0.206)
Random Forest (Brier=0.056)
SVM radial (Brier=0.056)
Bin size
30
C
PCA of digit-pattern feature space
Training clusters and target datasets
PC1 (21.7% variance)
RawData
ErrData
Target dataset
ErrData
RawData
Training data
Mixed-contamination
Simulated digit-preference irr
Simulated normal
D
Top feature importance
Integrated importance from machine-learning feature ranking
Feature
Integrated importance
Feature class
Joint digit
Progressive
Single digit
Importance
E
Confusion matrices of machine-learning models
Elastic-net Logistic
Ensemble
Isolation Forest
Random Forest
SVM radial
Actual class
Predicted class
Normal
Irregular
Count
F
Risk scores of target datasets
Risk score
Model
Elastic_net_Logistic
Ensemble_risk
Isolation_Forest
Random_Forest
SVM_radial
Dataset
ErrData
RawData
G
Isolation Forest anomaly score distribution
Target datasets are marked by vertical lines
Density
Isolation Forest anomaly risk
RawData 0.124
ErrData 0.302
Target dataset
ErrData
RawData
Training data
Mixed-contamination
Simulated digit-preference irregular
Simulated normal
H
Model performance summary
Model
Metric
SVM radial
Random Forest
Isolation Forest
Ensemble
Elastic-net Logistic
Accuracy
AUC
Balanced_accuracy
Brier_score
F1_score
Precision
Sensitivity
Specificity
Relative value
I
Full-sample statistical summary of target datasets
Joint decimal digits 3-4
Single decimal digit 3
Dataset
Metric
ChiSquare
CramersV
H_star
KL
Pref_nest_combo
Pref_same_pair
Pref_sequential_combo
Pvalue
Relative value
J
Isolation Forest anomaly risk
Training distributions and target datasets
Anomaly risk
ErrData 0.302
RawData 0.124
Target dataset
ErrData
RawData
Training data
Mixed-contamination
Simulated digit-preference irregular
Simulated normal
K
Progressive subsampling metrics of target datasets
-log10(P)
Chi-square
Cramér's V
KL divergence
Median P value
Normalized entropy H*
Metric value
Sample size (n)
Dataset
ErrData
RawData

Figure S1

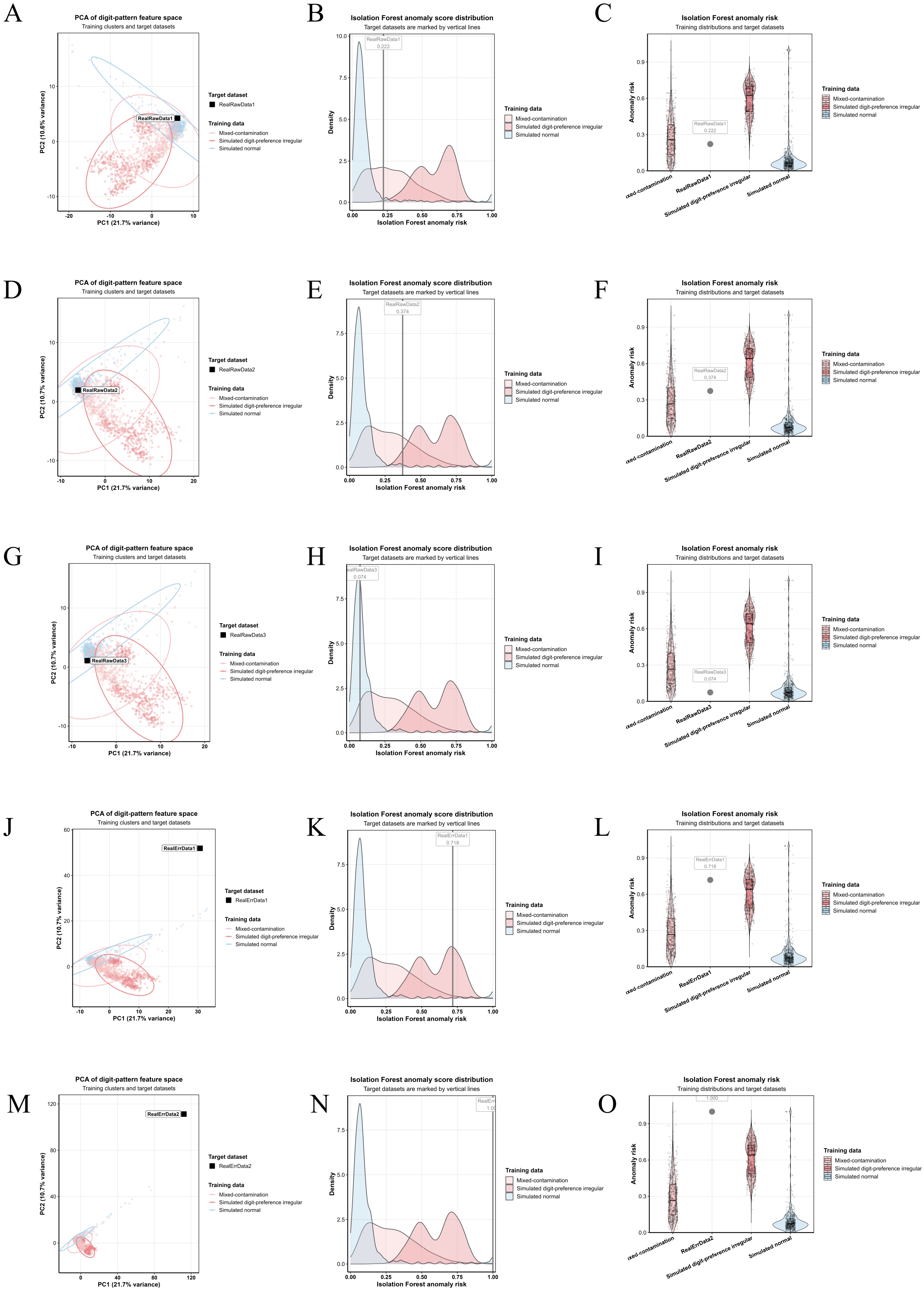

A
PCA of digit-pattern feature space
Training clusters and target datasets
PC1 (21.7% variance)
PC2 (10.6% variance)
RealRawData1
Target dataset
Training data
Mixed-contamination
Simulated digit-preference irregular
Simulated normal
B
Isolation Forest anomaly score distribution
Target datasets are marked by vertical lines
RealRawData1
0.222
Density
Isolation Forest anomaly risk
C
Isolation Forest anomaly risk
Training distributions and target datasets
Anomaly risk
D
PCA of digit-pattern feature space
Training clusters and target datasets
PC2 (10.7% variance)
RealRawData2
E
RealRawData2
0.374
F
G
RealRawData3
H
0.074
I
J
RealErrData1
K
0.718
L
M
RealErrData2
N
O
1.000

Figure S2

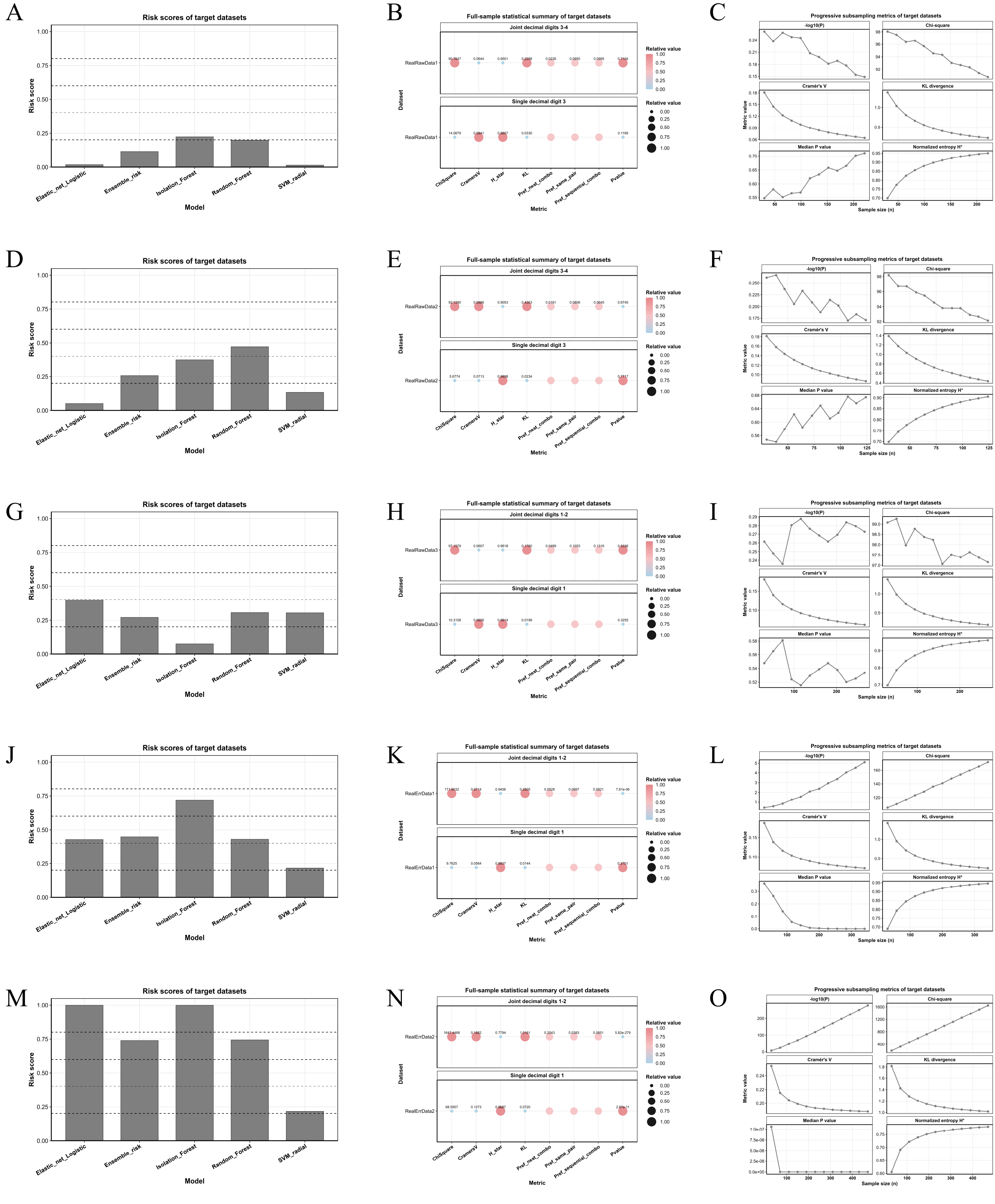
A
Risk scores of target datasets
Risk score
Model
Elastic_net_Logistic
Ensemble_risk
Isolation_Forest
Random_Forest
SVM_radial
B
Full-sample statistical summary of target datasets
Joint decimal digits 3-4
Single decimal digit 3
RealRawData1
Dataset
Metric
Relative value
ChiSquare
CramersV
H_star
KL
Pref_next_combo
Pref_same_pair
Pref_sequential_combo
Pvalue
C
Progressive subsampling metrics of target datasets
-log10(P)
Chi-square
Cramér's V
KL divergence
Median P value
Normalized entropy H*
Metric value
Sample size (n)
D
E
RealRawData2
F
G
H
Joint decimal digits 1-2
Single decimal digit 1
RealRawData3
I
J
K
RealErrData1
L
M
N
RealErrData2
O

Table 1

| Integrated risk score | Risk grade | Interpretation |
|---|---|---|
| < 0.20 | Grade 0 | No apparent digit-pattern irregularity |
| 0.20–0.40 | Grade 1 | Mild irregularity |
| 0.40–0.60 | Grade 2 | Moderate irregularity; further review recommended |
| 0.60–0.80 | Grade 3 | High irregularity; raw records should be checked |
| ≥ 0.80 | Grade 4 | Very high irregularity; detailed verification is recommended |

Table 2

| Dataset | n | Tested digit | $\chi^2$ | df | *P* value | Cramér's V | Normalized entropy ($H^*$) | Interpretation |
|---|---|---|---|---|---|---|---|---|
| RawData | 253 | 3rd decimal digit | 7.9091 | 9 | 0.5433 | 0.0589 | 0.9932 | No significant deviation |
| ErrData | 255 | 3rd decimal digit | 17.5882 | 9 | 0.0403 | 0.0875 | 0.9851 | Significant deviation |

Table 3

| Dataset | Sample size | Iterations | Mean $\chi^2$ | Median $P$ | Mean Cramér's V | Mean $H^*$ | Mean KL | Median $-\log_{10}P$ |
|---|---|---|---|---|---|---|---|---|
| RawData | 30 | 1000 | 97.6800 | 0.5476 | 0.1809 | 0.6993 | 1.3846 | 0.2615 |
| RawData | 152 | 1000 | 90.6250 | 0.7272 | 0.0775 | 0.9260 | 0.3409 | 0.1384 |
| RawData | 253 | 1 | 84.9447 | 0.8419 | 0.0582 | 0.9628 | 0.1715 | 0.0747 |
| ErrData | 30 | 1000 | 99.3200 | 0.5476 | 0.1825 | 0.6966 | 1.3972 | 0.2615 |
| ErrData | 153 | 1000 | 101.3830 | 0.4184 | 0.0817 | 0.9155 | 0.3891 | 0.3784 |
| ErrData | 255 | 1 | 103.8235 | 0.3503 | 0.0641 | 0.9485 | 0.2371 | 0.4556 |

Table 4

| Model | Threshold | Accuracy | Sensitivity | Specificity | Precision | NPV | F1 score | Balanced accuracy | AUC | Brier score |
|---|---|---|---|---|---|---|---|---|---|---|
| Random Forest | 0.491221 | 0.926667 | 0.91 | 0.96 | 0.978495 | 0.842105 | 0.943005 | 0.935 | 0.9809 | 0.055658 |
| Elastic-net Logistic | 0.494828 | 0.923333 | 0.915 | 0.94 | 0.968254 | 0.846847 | 0.940874 | 0.9275 | 0.98395 | 0.048439 |
| SVM radial | 0.550641 | 0.916667 | 0.9 | 0.95 | 0.972973 | 0.826087 | 0.935065 | 0.925 | 0.9754 | 0.056181 |
| Isolation Forest | 0.198629 | 0.863333 | 0.85 | 0.89 | 0.939227 | 0.747899 | 0.892388 | 0.87 | 0.8922 | 0.206379 |
| Ensemble | 0.422108 | 0.913333 | 0.91 | 0.92 | 0.957895 | 0.836364 | 0.933333 | 0.915 | 0.97895 | 0.065783 |

Table 5

| Dataset | n | Random Forest | Elastic-net Logistic | SVM radial | Isolation Forest | Ensemble risk | Risk grade |
|---|---|---|---|---|---|---|---|
| RawData | 253 | 0.170373 | 0.149674 | 0.054441 | 0.12402 | 0.124627 | Grade 0: no apparent irregularity |
| ErrData | 255 | 0.787406 | 0.950056 | 0.923895 | 0.301684 | 0.74076 | Grade 3: high irregularity |

Table 6

| Dataset | n | Random Forest | Elastic-net Logistic | SVM radial | Isolation Forest | Ensemble risk | Risk grade |
|---|---|---|---|---|---|---|---|
| RealRaw Data1 | 221 | 0.197956 | 0.017005 | 0.013691 | 0.221924 | 0.112644 | Grade 0: no apparent irregularity |
| RealRaw Data2 | 124 | 0.470774 | 0.049999 | 0.133819 | 0.373728 | 0.25708 | Grade 1: mild irregularity |
| RealRaw Data3 | 266 | 0.305814 | 0.399013 | 0.304079 | 0.074306 | 0.270803 | Grade 1: mild irregularity |
| RealErr Data1 | 341 | 0.430195 | 0.428039 | 0.215075 | 0.717983 | 0.447823 | Grade 2: moderate irregularity |
| RealErr Data2 | 470 | 0.74392 | 0.999978 | 0.215075 | 1 | 0.739743 | Grade 3: high irregularity |

# Table S1

| Dataset | Sample-size range | $\Delta\chi^2$ | Δmedian *P* | ΔCramér's V | Δ*H** | ΔKL | Overall trend |
|---|---|---|---|---|---|---|---|
| RawData | 30–253 | −12.7353 | +0.2943 | −67.8% | +0.2634 | −87.6% | Deviations progressively attenuated |
| ErrData | 30–255 | +4.5035 | −0.1973 | −64.9% | +0.2519 | −83.0% | Deviation signal persisted relatively more strongly |

Table S2

| Dataset | Cramér's V slope | $H^*$ slope | KL slope | Median $P$ slope | $-\log_{10}P$ slope | $\chi^2$slope |
|---|---|---|---|---|---|---|
| RawData | −0.0552 | +0.1203 | −0.5539 | +0.1319 | −0.0838 | −5.9736 |
| ErrData | −0.0529 | +0.1141 | −0.5254 | −0.0870 | +0.0855 | +2.1594 |